# Towards Clinical AI Fairness: Filling Gaps in the Puzzle

Mingxuan Liu[1#], Yilin Ning[1#], Salinelat Teixayavong[1], Xiaoxuan Liu[2,3,4], Mayli Mertens[5,6], Yuqing Shang[1], Xin Li[1], Di Miao[1], Jie Xu[7], Daniel Shu Wei Ting[1,8], Lionel Tim-Ee Cheng[9], Jasmine Chiat Ling Ong[10], Zhen Ling Teo[8], Ting Fang Tan[8], Narrendar RaviChandran[8], Fei Wang[11], Leo Anthony Celi[12,13,14], Marcus Eng Hock Ong[15,16], Nan Liu[1,15,17]*

[1] Centre for Quantitative Medicine, Duke-NUS Medical School, Singapore, Singapore
[2] College of Medical and Dental Sciences, University of Birmingham, Birmingham, UK
[3] University Hospitals Birmingham NHS Foundation Trust, Birmingham, UK
[4] National Institute for Health and Care Research Birmingham Biomedical Research Centre, University of Birmingham, Birmingham, UK
[5] Centre for Ethics, Department of Philosophy, University of Antwerp, Antwerp, Belgium
[6] Antwerp Center on Responsible AI, University of Antwerp, Antwerp, Belgium
[7] Department of Health Outcomes and Biomedical Informatics, University of Florida, Gainesville, FL, USA
[8] Singapore Eye Research Institute, Singapore National Eye Centre, Singapore, Singapore
[9] Department of Diagnostic Radiology, Singapore General Hospital, Singapore, Singapore
[10] Department of Pharmacy, Singapore General Hospital, Singapore, Singapore
[11] Department of Population Health Sciences, Weill Cornell Medicine, New York, NY, USA
[12] Laboratory for Computational Physiology, Massachusetts Institute of Technology, Cambridge, MA, USA
[13] Division of Pulmonary, Critical Care and Sleep Medicine, Beth Israel Deaconess Medical Center, Boston, MA, USA
[14] Department of Biostatistics, Harvard T.H. Chan School of Public Health, Boston, MA, USA
[15] Programme in Health Services and Systems Research, Duke-NUS Medical School, Singapore, Singapore
[16] Department of Emergency Medicine, Singapore General Hospital, Singapore, Singapore
[17] Institute of Data Science, National University of Singapore, Singapore, Singapore

[#] These authors contributed equally
* Corresponding author: Nan Liu, Centre for Quantitative Medicine, Duke-NUS Medical School, 8 College Road, Singapore 169857, Singapore. Phone: +65 6601 6503. Email: liu.nan@duke-nus.edu.sg

## Abstract

The ethical integration of Artificial Intelligence (AI) in healthcare necessitates addressing fairness—a concept that is highly context-specific across medical fields. Extensive studies have been conducted to expand the technical components of AI fairness, while tremendous calls for AI fairness have been raised from healthcare. Despite this, a significant disconnect persists between technical advancements and their practical clinical applications, resulting in a lack of contextualized discussion of AI fairness in clinical settings. Through a detailed evidence gap analysis, our review systematically pinpoints several deficiencies concerning both healthcare data and the provided AI fairness solutions. We highlight the scarcity of research on AI fairness in many medical domains where AI technology is increasingly utilized. Additionally, our analysis highlights a substantial reliance on group fairness, aiming to ensure equality among demographic groups from a macro healthcare system perspective; in contrast, individual fairness, focusing on equity at a more granular level, is frequently overlooked. To bridge these gaps, our review advances actionable strategies for both the healthcare and AI research communities. Beyond applying existing AI fairness methods in healthcare, we further emphasize the importance of involving healthcare professionals to refine AI fairness concepts and methods to ensure contextually relevant and ethically sound AI applications in healthcare.

# Main

The desire to ensure fairness in artificial intelligence (AI) in high-stakes fields like healthcare has surged as a paramount ethical concern, garnering considerable attention in recent years.[1-5] In-depth studies have shed light on the extensive adoption of AI techniques across diverse medical fields, while also identifying fairness as a significant challenge impeding ethical AI integration in healthcare contexts.[6-11] AI biases, i.e., opposed to the principles of fairness, arise when attributes like age, gender, race, or socio-economic status unjustly impact model-based decision-making.[12,13] Health equity—the principle of providing equal opportunity for all human beings to attain their full health potential regardless of societal barriers[14,15], is jeopardized by such biases. With a lack of a commitment to fairness, AI applications can potentially exacerbate, rather than diminish, health inequalities.[16,17]

The context-specific nature of fairness in healthcare adds complexity to developing fair algorithm solutions.[17,18] AI fairness and its decision-making consequences differ markedly across medical fields, precluding a one-size-fits-all solution. First, bias-inducing variables (referred to as relevant variables), extending beyond frequently recorded sensitive variables such as age, gender/sex, and race/ethnicity, vary across medical fields and their bias-inducing mechanisms. For instance, dermatology encounters bias from data underrepresentation regarding skin tone—a relevant variable that can confound diagnostic models[19]. In contrast, gastroenterology and hepatology face different challenges. Particularly in liver transplantation, bias can arise from overlooked gender differences in commonly used clinical predictors despite existing evidence. For example, the Model for End-Stage Liver Disease (MELD) algorithm's reliance on creatinine often underestimates renal dysfunction in women, resulting in lower MELD scores for women compared to men of comparable disease severity.[7] Additionally, each medical field is represented by different types of data. For instance, radiology and pathology predominantly deal with imaging data, which is presumably more objective for decision-making but can still contain hidden biases[20]. In contrast, mental health often depends on self-reported outcomes and behavioral data, which are more subjective and susceptible to human cognitive bias from both patients and healthcare providers[8,21]. Fairness definitions also vary, reflecting the complex interplay between relevant variables and the outcomes of interest. In radiology, fairness emphasizes consistent model performance (e.g., in terms of equal false positive rates) across demographic subgroups, despite varied biological detection challenges (e.g., between male and female)[22,23]. In liver transplantation, fairness is guided by medical urgency over socioeconomic status to guarantee equitable treatment access and consistent decision-making (e.g., in terms of equal decision rates) across all subgroups[24].

While AI fairness and biases have been extensively studied[25-28], there remains a significant disconnect between these proposed solutions and the calls from healthcare. Our study goes beyond simply describing AI fairness components and narratively identifying several gaps; it sets itself apart from existing reviews by critically pinpointing the in-depth discrepancies between AI fairness techniques and their healthcare integration, employing evidence maps to offer straightforward data evidence regarding these gaps. Through a meticulous review of pertinent literature, we identify critical research gaps and application shortcomings in clinical AI fairness, thus shedding light on areas ripe for joint exploration by both healthcare professionals and AI researchers, ideally in collaboration with ethicists. Furthermore, we outline actionable steps to advance clinical AI fairness, aiming to bridge the gap and catalyze progress toward just, fair, and equitable healthcare delivery.

# Results

We conducted a systematic scoping review to analyze the current landscape of AI fairness in healthcare. Our search of five databases yielded 11,719 papers, of which 272 were included for analysis. Supplementary Figure S1 illustrates the selection procedure in detail. Our evidence map analysis uncovered critical deficiencies in clinical AI fairness, specifically in the applications of AI fairness across various medical specialties, the relevant variables considered, and the clinical datasets employed, which constitute the landscape of AI fairness in healthcare. Upon this, we pinpointed gaps in the methodologies used to identify, evaluate, and mitigate biases.

## Data in clinical AI fairness research

Figure 1 summarizes the distribution of studies across various medical fields, identifying predominant areas such as health informatics and policy, cancer, radiology, cardiology, infectious disease, mental health, critical care, neurology, and pulmonology. It also classifies studies based on the types of data utilized, the relevant variables (i.e., bias-inducing variables), and the employment of publicly accessible datasets. All studies reviewed were based on retrospective data. We found limited studies related to AI fairness in several medical fields, including otolaryngology, renal, family medicine, rheumatology, anesthesiology, hematology, oral health, immunology, occupational health, physical medicine, and rehabilitation.

In the reviewed studies, tabular static data emerged as the most frequently used data type (n=155, 57.0%), prevalent in various medical fields, whereas image data, ranking second in usage (n=66, 24.3%), was primarily used in specialized areas such as cancer, radiology, and dermatology. Tabular temporal data was occasionally used in research across 16 medical fields such as cancer and public health; each of these fields featured a maximum of five studies employing this data type. Other data types, such as video, signal, audio, and text, were used less frequently in the studies reviewed. Among all medical fields, the analysis of AI fairness in mental health exhibited the broadest diversity in data types, utilizing all seven categories, predominantly tabular static and text data. Conversely, AI fairness studies in dermatology mainly concentrated on image data.

The most prevalent relevant variables considered in AI fairness studies were gender/sex (n=153, 56.3%) and ethnicity/race (n=131, 48.2%), followed by age (n=85, 31.3%). These concept pairs – gender and sex, ethnicity and race – often lacked precision in reporting by the authors, so they were collectively summarized. Other than the three most commonly examined relevant variables, socioeconomic status was featured prominently in studies, followed by location, skin tone, health status, institute (e.g., medical center, hospital), and language literacy. Additional variables like education, anthropometry (weight and height), the experience level of clinicians, time dynamics, marital status, etc., were considered for AI fairness in at least two papers each. The category of “Others” included ten papers that dealt with relevant variables such as religion, sexual orientation, etc., each only appearing once in this review. Twenty-five papers did not specify relevant variables in AI fairness investigation.

Among the 272 included papers, most papers (n=181, 66.5%) analyzed publicly available datasets to implement AI fairness strategies (see Figure 1). We summarized these 152 publicly accessible datasets, grouped by data type and ordered by frequency. Many public datasets were only used once or twice in this review (n=115, see Supplementary Table S1). Table 1 displays the datasets that were employed more frequently, covering four data types: tabular static, tabular temporal, image, and text data.

## Methods in clinical AI fairness research

### Bias identification

Bias identification is pivotal in developing AI fairness techniques. Among the included papers, a majority of papers (n=163, 59.9%) positioned bias identification as a precursor to proposing bias-mitigation methods. In the reviewed studies, bias identification typically relied on literature evidence, exploratory data analysis, and method comparisons. Historical bias often included the overreliance on certain relevant variables or biases presented in established models[29,30]. Researchers frequently sought to uncover bias in data by employing statistical methods or visual tools to highlight disparities in data distributions that could suggest the presence of bias.[31,32] Data bias was frequently related to issues such as class imbalances and the underrepresentation of minorities.[31,33] Algorithmic bias was mainly detected by examining the model’s ability to discern relevant variables[34,35] or through the application of fairness metrics (see subsection “Measurements for bias-evaluation”), which were then compared across algorithms. For foundation models, bias identification focused on the design of prompts in medical contexts and their effects on outputs, such as on the representation of demographic realities.[36,37]

The rest of the papers (n=109, 40.1%) solely focused on bias identification, without providing bias-mitigation solutions. These studies typically employed regression models to detect data or real-world biases by examining the association of the involved relevant variables with the outcomes of interest.[38,39] Notably, some papers might implicitly assume that the models for detecting bias were accurate, attributing any detected biases to systemic issues. Some studies identified algorithm bias by comparing model performance between subgroups and considered performance variations as evidence for bias.[40,41] Despite these efforts, the underlying causes and mechanisms of biases usually remained elusive, often due to the lack of detailed analysis of data bias against real-world demographics[37] or established fairness standards.

**Bias evaluation**

Quantitative bias evaluation is crucial to detecting algorithmic bias and to assessing the efficacy of bias mitigation methods if applied (details explained in Box 1c). As shown in Figure 2, group fairness was the most common fairness notion (n=248, 91.2%), while individual fairness (n=17, 6.3%) and distribution fairness (n=13, 4.8%) were mentioned less often (see Box 1c and Figure 2 for more details). In this review, clinician involvement in fairness evaluations was observed, with some participating in experiments that compared clinicians' decision-making with and without AI models,[42] or assessing the alignment between model outputs and clinicians' own decision-making.[43,44] Thirty-nine papers applied explainable artificial intelligence (XAI) to advance fairness assessments, mainly by investigating the contribution of variables to prediction outcomes. This involved visualizing relevant variables' contribution towards predictions[45], comparing variable contribution across subpopulations[46,47], etc.

Group fairness was dominated by performance-based (n=184, 67.6%) and parity-based (n=84, 30.8%) metrics. These two types of metrics were jointly utilized in 38 papers. Equal opportunity and equalized odds[48] were the most popular metrics among the performance-based, and statistical (demographic) parity[49] was most common in the parity-based category. Performance-based metrics typically gauge disparities in model performance such as accuracy, true positive rate and false positive rate, yet calibration, a more granular measure of predictive performance, was infrequently factored into assessments (n=9). Rank-based metrics, focusing on continuous predictions, were used in 11 papers, to ensure that the severity of outcomes was consistently assessed across subgroups[50]. These metrics were primarily employed for clinical predictions such as length of hospital stays and patient survival. Remove-based metrics that focused on removing confounders or disentangling bias-related information were discussed in another seven papers.

For studies on individual fairness, five papers causally addressed individual fairness with counterfactual-based metrics, emphasizing the invariance of predictions against artificial changes in variables (e.g., change of a female identity into a male)[51-55]. Seven papers utilized similarity-based metrics, expecting comparable model outputs of similar individuals[56-62]. The calculation of individual similarity did not necessarily require specific relevant variables (as indicated in Figure 2, in the "individual fairness" part).

For studies on distribution fairness, the allocation of healthcare resources also did not necessarily involve certain relevant variables (as indicated in Figure 2, in the "distributive fairness" part). Participants in distributive fairness schemes could be both individuals or groups that are usually geographically apart (for example, patients from different hospitals). Among these studies, two types of objectives prevailed: to distribute equally among the participants, where variance-based metrics were applied (n=8), or to distribute appropriately based on the efforts and inputs of participants, where the reward-based metrics were applied (n=6). In this review, the adoption of federated learning (FL)—a framework that allows for cross-institute collaboration while preserving data privacy, was closely related to distributive fairness.[63-67] This approach could be combined with the technology of blockchain to enhance privacy and fairness (n=4).[63,64]

**Bias mitigation**

Out of the 163 papers that attempted to mitigate bias, the most utilized method was in-process (n=121, 74.2%), followed by pre-process (n=44, 27.0%) and post-process (n=15, 9.2%) methods. This indicates constructing inherently fair models (in-process; see Box 1a for detailed definition) was favored over adjusting data beforehand (pre-process) or correcting model outputs afterward (post-process). Some papers adopt multiple types of methods (n=14, 8.6%), which included studies that compared multiple types of existing methods (n=6) and studies that crafted fairness methods to mitigate bias across multiple stages (n=8).

Pre-process methods often involved resampling (such as under-sampling[31], over-sampling[68], stratified batch sampling[33], and fairness-enhanced sampling[69]), as well as reweighing[32,70,71] strategies to balance data distribution between subgroups or align it with population demographics. In-process methods often integrated fairness as model constraints[32,70,72], or applied representation learning (often via adversarial learning[73,74]) to address fairness by filtering bias-related information when modeling. Post-process methods usually included optimizing prediction thresholds[74,75], utilizing established methods like (calibrated) equalized odds post-processing[71,76], or finetuning model subgroup-wise[75]. However, addressing fairness during model implementation, or constructing post-modeling clinical pathways to mitigate bias were rare in this review[42]. Five papers engaged domain experts in model development to mitigate bias, such as embedding clinicians' expertise into model construction[44,77] and actively seeking their perspectives on potential sources of errors and (relevant) variable selection for the models[76,78,79].

# Discussion

In this review, we summarized pivotal advancements and highlighted gaps regarding healthcare data and methodologies that impede progress in clinical AI fairness. These gaps present opportunities that necessitate interdisciplinary efforts in both technological development and clinical involvement. We outline fundamental challenges, propose corresponding actionable strategies, and suggest future directions, as summarized in Table 2 and described below.

## Lack of AI fairness research across medical fields

We observed an absence of AI fairness research within various medical fields, as shown in Figure 1, despite calls for closer investigation of AI fairness in these fields[9-11]. Comparing research volume across medical fields may seem trivial, but fairness remains essential regardless of a field's prominence. The absence or low values indicates the overlooked AI fairness concerns in fields like anesthesiology and oral health, given the prevalence of AI research in these fields as documented in the literature, as summarized in Supplementary Table S2. For example, Morch et al.[11] identified around 200 studies about AI in oral health (up to 2020), while we only identified one study about AI fairness in oral health in this review. Even in fields with relatively higher AI fairness research volumes, such as cancer and cardiology, the attention to AI fairness is still insufficient compared to the extensive AI-related studies. As AI methodologies and applications evolve, integrating fairness considerations from the outset is crucial, rather than treating them as an afterthought.

In addition, Figure 1 shows a reliance on public databases in fields like medical informatics and health policy, oncology, cardiology, and radiology, and critical care. Yet, our review finds sparse use of many public datasets listed in Table S1, often single-study applications, suggesting a deficiency in standardized fairness benchmarking concerning various clinical questions. Moreover, the infrequent use of established databases like the National Health and Nutrition Examination Survey (NHANES) for AI fairness studies indicates that the potential utility of these resources may be under-valued or that they may not be sufficient due to the absence of bias-related information[80].

Addressing the challenges necessitates a multifaceted strategy. To tailor AI fairness for specific medical fields, it is crucial to summarize[81,82] and examine[83] field-specific datasets with the lens of fairness and collect new datasets with fairness concerns when existing data is insufficient. Concerted efforts are needed to collect data with rigorous strategies that ensure data diversity and representativeness[84], with clinician-assisted curation to mitigate potential data bias where necessary[19,80]. Importantly, such efforts should include improvements to medical devices to generate data with equal quality from subpopulations (e.g., when measuring blood oxygen for people with varying skin tone), and the use of less biased proxies for health-related measures (e.g., cystatin C as a gender-neutral alternative to creatinine for measuring renal function[85]). Additionally, making medical data publicly available, with attention to both access and privacy[86], will foster thorough AI fairness research in compliance with HIPAA standards.

## Beyond sensitive variables

While the examination of gender/sex, race/ethnicity, and age is prevalent in AI fairness, it does not encompass the full spectrum of concerns as bias can arise from a wide range of variables (i.e., relevant variables). For example, as previously mentioned, even non-sensitive parameters such as creatinine may inadvertently induce biases through the setting of reference ranges that are not adequately representative.[7] On the other hand, for the commonly mentioned sensitive variables, clarification is needed to resolve the confusion in concepts. For example, "gender" as a social identity was often conflated with "sex" as a biological concept—a discrepancy that needs clear differentiation to enhance the validity of fairness assessments.[18]

Variables of health status, skin tone, institute, language literacy, marital status, anthropometry (weight and height), education level, and experience level of clinicians are acknowledged as influential yet less frequently considered in AI fairness. Most of these relevant variables are patient-oriented, while characteristics of healthcare providers can also induce biases (e.g., via a patient-physician gender discordance[87]) but are seldom available for modeling. These attributes, though not all traditionally classified as sensitive, can introduce biases into data-driven decision-making. In addition, the intersectional effects of these variables were relatively rarely considered. Accurately acquiring and reporting these variables is challenging but necessary for close bias investigation.

Moreover, many papers from the technical communities utilize certain datasets and bias-inducing variables to demonstrate their bias mitigation techniques, without providing a clear justification for their handling of relevant variables. They also might not assess initial bias profiles beyond class imbalance and varying prevalence rates[84], let alone elucidate the clinical impact of bias-mitigation techniques. This necessitates further comprehension of clinical contexts and precise characterization of fairness scenarios before developing AI fairness methods.

## Addressing individual fairness

While group fairness metrics (parity-based and performance-based) were frequently researched, as indicated in Figure 2, there is a marked deficiency in individual fairness to ensure comparable individuals are treated equally. While policymakers and healthcare managers prioritize group fairness based on population ethics, individual fairness, which directly affects patients and is aligned with clinical ethics, is equally essential.[17,88] The rise in precision medicine amplifies the need for a balanced approach to AI fairness, incorporating both broad population insights and individual patient nuances to achieve equitable healthcare outcomes[89].

The lack of investigation on individual fairness in current literature may indicate the difficulty of integrating it into clinical AI fairness. We can take an intermediate step through an intersectional approach that considers multiple relevant variables for fairness considerations at a time to capture more accurate individual profiles. AI researchers can develop fairness metrics that can measure biases at a granular level, capturing individual variations and meticulously measuring individual similarity in clinical settings[72]. Subsequently, researchers can cater bias-mitigation methods to individual fairness, for example, crafting personalized treatment recommendations that are sensitive to individual variations in disease presentation and response to treatment while avoiding unjustified effects from relevant variables.[88] The downside is that incorporating additional personal attributes to enhance AI fairness can inadvertently increase the risk of compromising individual privacy. As it remains a subject of debate whether the distinction between individual and group fairness is merely a matter of granularity[90], it is beneficial to explore additional potential pathways that connect group fairness with individual fairness. Moreover, algorithms should be transparent in their decision-making, ensuring individuals can comprehend the impact on their outcomes (e.g., diagnoses and treatment recommendations)[72], which goes beyond broad group identities to recognize and respect individual patient autonomy.

## Investigating distribution fairness

The current research on clinical AI fairness also significantly overlooks distribution fairness, which is vital for the equitable distribution of physical resources, such as vaccines and treatments, and virtual resources, such as data capacity and model performance, across participants (individuals or groups). While a few studies in this review addressed distribution fairness, they mainly focused on computational aspects such as data collaboration and model co-training. Furthermore, these approaches put more emphasis on equal distribution among participants (variance-based) than the balance of contribution and benefit across participants (reward-based). Fair distribution involves creating fairness-aware models that address the allocation of physical medical services directly. Regarding virtual resources, fair distribution involves utilizing techniques such as federated learning with incentive mechanisms to rationally utilize data resources and fairly reward model performance across participants. Moreover, addressing bias across institutions and geographic locations is vital, as data variability across sources can impact model fairness.[74]

## Putting AI fairness into clinical contexts

Moreover, current AI fairness techniques lack contextual ties to healthcare scenarios. Group fairness, which concentrates on “outcome fairness” (i.e., the fairness of the decision outcomes)[91,92], makes up the majority of evaluation metrics. In-process methods also dominate the fairness methodologies, followed by post-process methods, which usually involve “black-box” models with obscure data bias and bias-mitigation paths. This obscures “procedural fairness” (i.e., the fairness in the decision-making process)[93,94]—another perspective of fairness that is largely neglected in the current literature. Even for approaches like reinforcement learning, which addresses the interaction of model development and context, the corresponding fairness methodology typically focuses only on the final outputs[95]. This also applies to large language (or vision) models that are based on reinforcement learning and encapsulated as interfaces for healthcare applications. Furthermore, rarely involving clinicians early in model development—beyond just evaluating the model outputs—misses opportunities to integrate clinical expertise into bias mitigation, which could strengthen the rationale behind model decisions.

Clinical AI fairness is not a one-time step, as biases can enter at any stage, affecting eventual patient outcomes.[17] To cultivate clinically contextualized fairness, it is essential to involve domain experts familiar with specific clinical contexts and aware of biases to participate in the model development and evaluation process actively. Despite some papers reviewed highlighting the role of domain experts in bias assessment and mitigation, more sensible ways of integrating domain experts need to be explored. This can include matching machine learning tasks with clinical problem formulation[96], causal modeling with clinically precise measurements[97] (e.g., using healthcare expenditures as a proxy for health needs is often implausible[98]), creating clinician-friendly module interfaces, formulating clinical guidelines for AI fairness applications, crafting interpretable AI fairness methods to enhance dialogue between all stakeholders, and so on. Moreover, in response to biased models, developing clinical pathways aimed at equitable health outcomes can provide a non-technological approach to ensure fairness in healthcare delivery. For instance, offering patients prone to underdiagnosis an additional test or an earlier evaluation by a clinician can help align model outcomes with real-world clinical decision-making.[1,5]

## Limitations

This review has several limitations. First, due to the ambiguous nature of terms like "bias" and "fair", our search strategy may have missed relevant studies that did not explicitly mention this concept in the title and abstract. Second, while our classification of medical fields and demographic constructs (such as race and ethnicity) aimed to be comprehensive, our taxonomy may not fully align with global classification systems due to international variations. Last but not least, by including only English-language papers, we may have limited the scope of our fairness discussion, potentially omitting insights on global health disparities.

## Conclusion

The disconnect between AI fairness techniques and the urgent demands in healthcare fields is evident through the gaps in clinical applications and the development of AI fairness methods. Joint exploration by healthcare professionals and AI researchers, preferably assisted by ethicists, is essential to expand beyond traditional sensitive variables, refine fairness concepts, and contextualize AI fairness in healthcare scenarios, thereby bridging current gaps. The pursuit of just, equitable, and fair healthcare delivery is advancing, propelled by interdisciplinary partnerships and a shared commitment to excellence in patient care and technological innovation.

## Methods

### Search strategies

This review followed the Preferred Reporting Items for Systematic reviews and Meta-Analyses extension for Scoping Reviews (PRISMA-ScR) checklist[99]. We searched five databases (MEDLINE, Web of Science, Embase, IEEE Xplore, ACM library) up to December 21, 2023. Detailed search strategies are provided in Supplementary Table S3. Six reviewers (ML, ST, YN, YS, XL, and DM) screened the articles based on the eligibility criteria. We randomly split papers into three non-overlapped groups. A pair of reviewers (ML with XL, YN with DM, and ST with YS) was allocated to each group with a similar number of publications. These reviewers individually screened the respective papers. Any disagreements were resolved through discussions among the six reviewers, under the oversight of a supervising reviewer (NL).

### Exclusion criteria

Papers with any of the following reasons were excluded: the paper was not in the medical or clinical domain, the paper was not about fair AI, i.e., the included study should address bias and fairness in AI techniques with a clear definition or quantification of fairness, the paper was not published as a research article (e.g., conference poster, conference abstract, or book chapter), the paper was a review paper, or the paper was not written in English.

### Data extraction and analysis

We extracted information regarding three main aspects: a) the healthcare context for AI fairness research (including the healthcare datasets, data types, and involved clinical specialty); b) the techniques of AI fairness developed or applied in healthcare applications (including fairness definition, bias identification, bias mitigation, and fairness evaluation), and c) techniques that can improve AI fairness (including explainable artificial intelligence and federated learning framework).

To clarify the topic of fairness, the grouping of fairness methods was organized according to Mehrabi et al.[91] and Xu et al.[100], as shown in Box 1a. We classified fairness definition into three categories according to Balakrishnan et al.[93]: group fairness, individual fairness, and distribution fairness, as detailed in Box 1b. We categorized existing fairness evaluation metrics into these three definitions, as indicated in Box 1c.

To visually report the results, descriptive summaries utilizing counts and proportions and figures such as evidence maps with marginal bar plots were reported, along with narrative interpretation when appropriate. Python version 3.8.3 (Python Software Foundation, Delaware, USA) and R version 4.0.2 (The R Foundation for Statistical Computing) were used for data analysis.

## Acknowledgments

We thank Dr. Stephen Pfohl for his suggestions on improving this paper. This work was supported by the Duke-NUS Signature Research Programme funded by the Ministry of Health, Singapore. Any opinions, findings and conclusions or recommendations expressed in this material are those of the author(s) and do not reflect the views of the Ministry of Health. Y.N. was supported by the Khoo Postdoctoral Fellowship Award (project no. Duke-NUS-KPFA/2021/0051) from the Estate of Tan Sri Khoo Teck Puat. M.M. is funded by the European Union, through the HORIZON-MSCA-2022-PF-01-01 Marie Curie Postdoctoral Fellowship (project 101107292 ‘PredicGenX’).

## Competing Interests

The authors declare that there are no competing interests.

## Author Contributions

M.L. and Y.N. contributed equally. Initial development of ideas: M.L., Y.N., S.T., N.L. Drafting of the manuscript: M.L. and Y.N. Critical revision of the manuscript: M.L., Y.N., X.L., N.L. Interpretation of the content: all authors. Revisions of the manuscript: all authors. Final approval of the completed version: all authors. Overseeing the project: N.L.

**Box 1. Term box**

a) Method to mitigate bias and strive for fairness

| Methods category | Description | Prototype methods |
|---|---|---|
| Pre-process | To remove bias in the model development data | Resampling, reweighting |
| In-process | To build a model with properties which are intended to optimize fairness and/or reduce bias, with original data that may or may not have bias | Representation learning (e.g., adversarial learning), regularization, subgroup modelling |
| Post-process | To modify the existed model according to different subpopulations before and during the implementation | (Calibrated) Equalized odds post-processing, threshold optimization for subgroups, fine-tune the model according to subgroups |

b) Fairness definitions

| Name | Description |
|---|---|
| Group fairness | To dictate an equal outcome probability between subgroups separated by a relevant variable |
| Individual fairness | To encode the notion that comparable individuals should be treated equally |
| Distribution fairness | To address fair distribution of limited resources (e.g., vaccination) across multiple stakeholders that can be both individuals or groups |

c) Metrics of fairness evaluation

| Fairness definition | Metrics category | Description | Prototype metrics |
|---|---|---|---|
| Group fairness | Parity-based | Focusing on predicted positive values | Demographic (statistical) parity, disparate impact, conditional statistical parity |
| | Performance-based | 1) Addressing the equality of performance metrics (e.g., accuracy, sensitivity [TPR], specificity [TNR], etc.) among different subgroups<br>2) Focusing on the calibration between average predicted probability and fractions of positive values | Equal opportunity, equalized odds, other metrics computed by differences or ratios of machine learning metrics among subgroups |
| | Rank-based | Focusing on the relative ranking of scores among outcome classes (e.g., mortality and survival), expected to be independent on group identity regarding relevant variables | Disparity in bipartite-ranking metrics (e.g., ΔxAUC) |
| | Remove-based | Focusing on the removal of bias-related information or confounders | Under blindness, Mutual information |
| Individual fairness | Similarity-based | Emphasizing similar results from similar individuals | Fairness with awareness |
| | Counterfactual-based | Emphasizing unchanged results after changing the sensitive information | Counterfactual fairness |
| Distribution fairness | Variance-based | Emphasizing the equality of quantities received by participants via minimizing variation | Variance or standardized deviation of the quantities (e.g., accuracy, loss, etc.) |
| | Reward-based | Emphasizing alignments between the quantities received by participants and their inputs and efforts | Reward based on the size of dataset |

**Table 1.** Popular public datasets for AI fairness investigation

| Data type | Public dataset | Medical fields | Relevant variables | Number of papers |
|---|---|---|---|---|
| Tabular Static | MIMIC-III, MIMIC-IV | Cardiology, CC, ED, Geriatrics, Informatics & Policy, Pharmacy, Renal | Age, Ethnicity/Race, Gender/Sex, Language, Marital status, Socioeconomics, Not Specified | 16 |
| | MEPS | Informatics & Policy, Oral Health, Public | Age, Ethnicity/Race, Gender/Sex, Socioeconomics, Not Specified | 11 |
| | Heritage Health | Informatics & Policy | Age, Gender/Sex, Not Specified | 8 |
| | SEER | Cancer, Gastroenterology, Hepatology, Pathology, Pulmonary | Age, Ethnicity/Race, Gender/Sex, Health status, Socioeconomics | 5 |
| | eICU | CC, Geriatrics | Age, Ethnicity/Race, Gender/Sex, Institute, Language | 4 |
| | Heart Disease | Cardiology | Age | 3 |
| | SUPPORT | Cancer, Cardiology, CC, Gastroenterology, Hepatology, ID, Neurology, Pulmonary | Age, Ethnicity/Race, Gender/Sex | 3 |
| Tabular Temporal | MIMIC-III, MIMIC-IV | Cancer, CC, ID, Informatics & Policy, Pulmonary | Age, Ethnicity/Race | 3 |
| Image | MIMIC-CXR | Cardiology, Dermatology, Pulmonary, Radiology | Age, Ethnicity/Race, Gender/Sex, Skin tone, Socioeconomics | 6 |
| | CheXpert | Cardiology, Dermatology, Pulmonary, Radiology | Age, Ethnicity/Race, Gender/Sex, Skin tone | 5 |
| | Chest-Xray8, Chest-Xray14 | Cardiology, Pulmonary, Radiology | Age, Ethnicity/Race, Gender/Sex | 5 |
| | Fitzpatrick17k | Cancer, Dermatology | Age, Gender/Sex, Health status, Skin tone | 5 |
| | ISIC | Cancer, Dermatology, Radiology | Gender/Sex, Skin tone | 5 |
| | ABIDE | Mental, Neurology, Radiology | Age, Ethnicity/Race, Gender/Sex, Health status, Others | 3 |
| | EyePACS | Endocrinology, Eye | Skin tone, Not Specified | 3 |
| | TCGA | Cancer, Pathodology, Pathology, Pulmonary, Renal | Ethnicity/Race, Institute, Not Specified | 3 |
| | UK Biobank CMR images | Cardiology, Radiology | Ethnicity/Race, Gender/Sex | 3 |
| Text | Twitter | Mental, O & G | Ethnicity/Race, Location | 3 |

[1]refers to those medical fields where the databases were applied as observed in this review, rather than how these databases are defined.
The complete list of public datasets used for AI fairness in this review can be found in Supplementary Table S1.

**Table 2.** Challenges of clinical AI fairness and how they may be overcome with future development

| Challenges | Description | Strategies |
|---|---|---|
| Data: insufficient medical domain-specific data | Insufficient availability of medical domain-specific datasets. | - Examine and summarize well-established databases for AI fairness investigation<br>- Collect medical field-specific datasets<br>- Publicize medical data with privacy protection |
| Data: Beyond "sensitive" variables | Inadequate examination of different factors that can introduce biases. | - Actively identify and discuss factors that may induce biases in healthcare data<br>- Be cautious about the terminology used |
| Technique: individual fairness oversight | Limited research on fairness at the individual level. | - Consider more individual-specific factors (e.g., language literacy)<br>- Develop bias-mitigation metrics and methods specifically designed for individual fairness. |
| Technique: distributive fairness neglect | Insufficient exploration of fairness issues in resource allocation. | - For physical healthcare resource allocation, develop fairness-aware models and mechanisms.<br>- For computational resources, address fairness through techniques like federated learning with incentive mechanisms. |
| Overall: lack of contextualization | Lack ties to healthcare scenarios | - Involve clinicians in the early stages of modeling<br>- Develop interpretable AI fairness methods.<br>- Establish clinical guidelines for the usage of AI fairness metrics. |

**Figure 1.** The evidence gap analysis of AI fairness methodology developments and applications in cross-tabulation between medical fields and data types, relevant variables, and public dataset utilization

| Medical field | Total | Tabular Static | Tabular Temporal | Image | Video | Signal | Audio | Text | Ethnicity/Race | Gender/Sex | Age | Socioeconomics | Location | Skin tone | Health status | Institute | Language | Education | Anthropometry | Clinician expertise | Time | Marital status | Others | Not Specified | Public data only | Used own data |
|---|---|---|---|---|---|---|---|---|---|---|---|---|---|---|---|---|---|---|---|---|---|---|---|---|---|---|
| Total | | 155 | 26 | 66 | 6 | 11 | 4 | 19 | 153 | 131 | 85 | 34 | 17 | 14 | 11 | 11 | 10 | 7 | 3 | 3 | 3 | 2 | 10 | 21 | 181 | 91 |
| Informatics & Policy | 51 | 42 | 4 | 0 | 1 | 1 | 0 | 4 | 30 | 20 | 16 | 4 | 4 | 1 | 0 | 0 | 1 | 1 | 0 | 0 | 1 | 0 | 2 | 6 | 39 | 12 |
| Cancer | 47 | 21 | 5 | 21 | 2 | 0 | 0 | 0 | 25 | 12 | 10 | 5 | 3 | 7 | 3 | 5 | 0 | 0 | 0 | 2 | 1 | 0 | 1 | 2 | 31 | 16 |
| Radiology | 36 | 8 | 0 | 34 | 0 | 0 | 0 | 0 | 20 | 23 | 18 | 4 | 2 | 1 | 2 | 1 | 1 | 1 | 1 | 0 | 0 | 0 | 3 | 1 | 26 | 10 |
| Cardiology | 34 | 21 | 2 | 8 | 0 | 4 | 0 | 0 | 23 | 25 | 15 | 5 | 1 | 0 | 0 | 0 | 1 | 1 | 0 | 0 | 0 | 0 | 0 | 0 | 24 | 10 |
| ID | 30 | 20 | 3 | 6 | 0 | 0 | 1 | 1 | 17 | 17 | 13 | 1 | 4 | 0 | 2 | 3 | 1 | 0 | 0 | 0 | 1 | 0 | 0 | 2 | 18 | 12 |
| Mental | 29 | 13 | 3 | 2 | 2 | 2 | 2 | 6 | 16 | 18 | 9 | 3 | 2 | 0 | 1 | 1 | 2 | 1 | 0 | 0 | 0 | 0 | 3 | 0 | 14 | 15 |
| CC | 26 | 21 | 3 | 0 | 0 | 0 | 0 | 3 | 19 | 17 | 7 | 5 | 0 | 0 | 1 | 1 | 2 | 0 | 0 | 0 | 0 | 2 | 0 | 3 | 23 | 3 |
| Neurology | 25 | 7 | 1 | 12 | 1 | 3 | 2 | 1 | 10 | 14 | 8 | 1 | 1 | 0 | 1 | 2 | 1 | 0 | 0 | 0 | 0 | 0 | 1 | 4 | 16 | 9 |
| Pulmonology | 20 | 6 | 2 | 12 | 0 | 0 | 0 | 0 | 10 | 12 | 10 | 4 | 1 | 0 | 0 | 2 | 0 | 0 | 0 | 0 | 0 | 0 | 0 | 1 | 19 | 1 |
| Public | 19 | 11 | 5 | 0 | 0 | 0 | 1 | 2 | 10 | 10 | 9 | 7 | 1 | 0 | 1 | 0 | 1 | 2 | 0 | 0 | 0 | 0 | 1 | 2 | 11 | 8 |
| Pediatrics | 15 | 8 | 2 | 5 | 1 | 0 | 0 | 0 | 7 | 5 | 3 | 1 | 2 | 0 | 0 | 0 | 1 | 0 | 1 | 1 | 0 | 0 | 0 | 1 | 10 | 5 |
| Dermatology | 15 | 0 | 0 | 15 | 0 | 0 | 0 | 0 | 2 | 5 | 3 | 0 | 0 | 10 | 1 | 0 | 0 | 0 | 0 | 0 | 0 | 0 | 0 | 0 | 11 | 4 |
| ED | 13 | 9 | 1 | 0 | 0 | 1 | 0 | 2 | 9 | 7 | 5 | 1 | 2 | 0 | 0 | 2 | 1 | 0 | 0 | 0 | 0 | 0 | 0 | 0 | 5 | 8 |
| Endocrinology | 13 | 9 | 2 | 1 | 0 | 0 | 0 | 1 | 7 | 7 | 5 | 2 | 0 | 0 | 0 | 0 | 0 | 1 | 0 | 0 | 0 | 0 | 1 | 3 | 10 | 3 |
| O & G | 10 | 8 | 1 | 0 | 0 | 0 | 0 | 2 | 7 | 2 | 1 | 2 | 3 | 0 | 0 | 0 | 0 | 2 | 0 | 1 | 0 | 0 | 0 | 0 | 5 | 4 |
| Eye | 9 | 2 | 0 | 7 | 0 | 0 | 0 | 1 | 5 | 3 | 1 | 0 | 0 | 2 | 0 | 1 | 0 | 0 | 0 | 0 | 0 | 0 | 1 | 1 | 6 | 4 |
| Gastroenterology | 8 | 5 | 1 | 2 | 0 | 0 | 0 | 0 | 3 | 5 | 2 | 1 | 1 | 0 | 0 | 0 | 0 | 0 | 0 | 0 | 0 | 0 | 0 | 2 | 7 | 1 |
| Geriatrics | 7 | 4 | 1 | 2 | 0 | 0 | 0 | 0 | 4 | 5 | 6 | 1 | 1 | 0 | 2 | 0 | 0 | 0 | 0 | 0 | 0 | 0 | 0 | 0 | 5 | 2 |
| Hepatology | 7 | 7 | 0 | 0 | 0 | 0 | 0 | 0 | 4 | 7 | 3 | 1 | 0 | 0 | 0 | 0 | 0 | 0 | 0 | 0 | 0 | 0 | 0 | 0 | 7 | 0 |
| Pharmacy | 7 | 5 | 0 | 0 | 0 | 0 | 0 | 2 | 4 | 3 | 2 | 1 | 0 | 0 | 0 | 0 | 0 | 1 | 0 | 0 | 0 | 0 | 0 | 1 | 5 | 2 |
| Surgery | 7 | 4 | 0 | 1 | 2 | 0 | 0 | 0 | 3 | 3 | 1 | 1 | 1 | 0 | 1 | 0 | 1 | 0 | 0 | 2 | 0 | 0 | 0 | 1 | 1 | 6 |
| Orthopedics | 6 | 4 | 0 | 2 | 0 | 0 | 0 | 0 | 5 | 4 | 2 | 2 | 0 | 0 | 1 | 0 | 1 | 1 | 0 | 0 | 0 | 0 | 0 | 0 | 1 | 4 |
| Pathology | 5 | 2 | 0 | 4 | 0 | 0 | 0 | 0 | 3 | 0 | 0 | 0 | 0 | 0 | 0 | 1 | 0 | 0 | 0 | 0 | 0 | 0 | 0 | 2 | 6 | 0 |
| Urology | 5 | 2 | 1 | 2 | 0 | 0 | 0 | 0 | 4 | 1 | 1 | 1 | 0 | 0 | 0 | 1 | 0 | 0 | 0 | 0 | 1 | 0 | 0 | 0 | 2 | 3 |
| Otolaryngology | 3 | 1 | 0 | 3 | 0 | 0 | 0 | 0 | 2 | 2 | 1 | 0 | 1 | 0 | 1 | 0 | 0 | 0 | 0 | 0 | 0 | 0 | 1 | 0 | 2 | 1 |
| Family Medicine | 2 | 1 | 0 | 0 | 0 | 0 | 0 | 1 | 2 | 2 | 1 | 0 | 0 | 0 | 0 | 0 | 0 | 0 | 0 | 0 | 0 | 0 | 0 | 0 | 0 | 2 |
| Renal | 2 | 1 | 0 | 1 | 0 | 0 | 0 | 0 | 0 | 1 | 0 | 0 | 0 | 0 | 0 | 1 | 0 | 0 | 0 | 0 | 0 | 0 | 0 | 0 | 2 | 0 |
| Rheumatology | 2 | 1 | 0 | 1 | 0 | 0 | 0 | 0 | 1 | 2 | 1 | 0 | 0 | 0 | 0 | 0 | 0 | 0 | 1 | 0 | 0 | 0 | 0 | 0 | 2 | 0 |
| Anesthesiology | 1 | 1 | 0 | 0 | 0 | 0 | 0 | 0 | 1 | 1 | 1 | 1 | 0 | 0 | 1 | 0 | 1 | 0 | 0 | 0 | 0 | 0 | 0 | 0 | 0 | 1 |
| Hematology | 1 | 0 | 0 | 0 | 1 | 0 | 0 | 0 | 0 | 0 | 0 | 0 | 0 | 1 | 0 | 0 | 0 | 0 | 0 | 0 | 0 | 0 | 0 | 0 | 1 | 0 |
| Oral Health | 1 | 1 | 0 | 0 | 0 | 0 | 0 | 0 | 1 | 0 | 1 | 1 | 0 | 0 | 0 | 0 | 0 | 0 | 0 | 0 | 0 | 0 | 0 | 0 | 1 | 0 |
| Immunology | 0 | 0 | 0 | 0 | 0 | 0 | 0 | 0 | 0 | 0 | 0 | 0 | 0 | 0 | 0 | 0 | 0 | 0 | 0 | 0 | 0 | 0 | 0 | 0 | 0 | 0 |
| Occupational | 0 | 0 | 0 | 0 | 0 | 0 | 0 | 0 | 0 | 0 | 0 | 0 | 0 | 0 | 0 | 0 | 0 | 0 | 0 | 0 | 0 | 0 | 0 | 0 | 0 | 0 |
| PM & R | 0 | 0 | 0 | 0 | 0 | 0 | 0 | 0 | 0 | 0 | 0 | 0 | 0 | 0 | 0 | 0 | 0 | 0 | 0 | 0 | 0 | 0 | 0 | 0 | 0 | 0 |

**Figure 2.** The evidence gap analysis of fairness metrics in cross-tabulation between high-level categories based on the fairness notions (group fairness, individual fairness and distributive fairness) and relevant variables.

| Notion | Metric category | Ethnicity/Race | Gender/Sex | Age | Socioeconomics | Location | Skin tone | Health status | Institute | Language | Education | Anthropometry | Clinician expertise | Marital status | Time | Others | Not Specified | Total |
|---|---|---|---|---|---|---|---|---|---|---|---|---|---|---|---|---|---|---|
| **Group Fairness**<br>Expect equal results across subgroups | **Parity−Based**<br>Equal predicted positive rates across subgroups | 52 | 42 | 32 | 16 | 4 | 4 | 3 | 2 | 6 | 3 | 1 | 1 | 1 | 0 | 3 | 2 | 84 |
| | **Performance−Based**<br>Equal model performance across subgroups | 111 | 97 | 61 | 21 | 12 | 14 | 10 | 6 | 7 | 6 | 2 | 2 | 2 | 3 | 7 | 2 | 184 |
| | **Rank−Based**<br>Consistent score−based rankings across subgroups | 9 | 5 | 3 | 3 | 0 | 0 | 1 | 0 | 0 | 1 | 0 | 0 | 0 | 0 | 1 | 0 | 11 |
| | **Remove−Based**<br>Remove sensitive information or bias−related confounders | 1 | 5 | 0 | 0 | 0 | 0 | 0 | 0 | 0 | 0 | 1 | 0 | 0 | 0 | 0 | 1 | 7 |
| | **Others−Based (Group)**<br>E.g., measuring differences in subgroup data profiles, etc. | 1 | 2 | 2 | 0 | 0 | 0 | 0 | 0 | 0 | 0 | 0 | 0 | 0 | 0 | 0 | 0 | 2 |
| **Individual Fairness**<br>Expect equal results for comparable individuals | **Counterfactual−Based**<br>Invariance of decision−making against changes in sensitive information | 3 | 3 | 3 | 1 | 0 | 0 | 0 | 0 | 0 | 0 | 0 | 0 | 0 | 0 | 1 | 1 | 5 |
| | **Similarity−Based**<br>Similar predictions for similar individuals | 1 | 2 | 2 | 0 | 1 | 0 | 0 | 0 | 0 | 0 | 0 | 0 | 0 | 0 | 0 | 4 | 7 |
| | **Others−Based (Individual)**<br>E.g., de−diversity in individual gains, etc. | 1 | 3 | 1 | 0 | 0 | 0 | 0 | 0 | 0 | 0 | 0 | 0 | 0 | 0 | 0 | 2 | 5 |
| **Distribution Fairness**<br>Expect fair distribution of limited resources | **Reward−Based**<br>Reward resources based on contribution | 0 | 0 | 0 | 0 | 0 | 0 | 0 | 2 | 0 | 0 | 0 | 0 | 0 | 0 | 0 | 5 | 6 |
| | **Variance−Based**<br>Equal resources received by participants | 0 | 0 | 0 | 0 | 2 | 0 | 0 | 3 | 0 | 0 | 0 | 0 | 0 | 0 | 0 | 4 | 8 |

# Supplementary Materials

## **Supplementary Figure S1.** PRISMA flow diagram

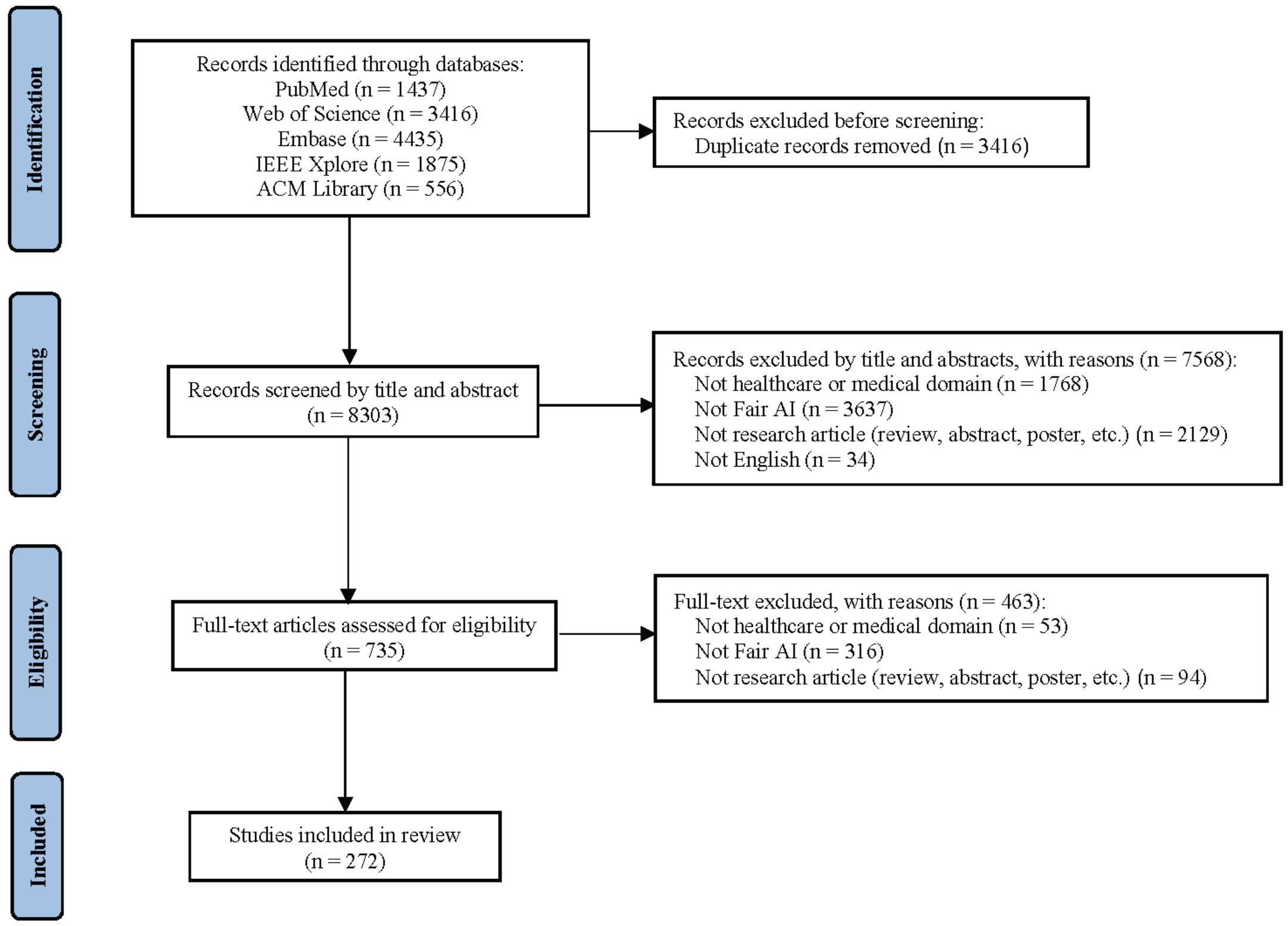

**Supplementary Table S1.** List of public datasets stratified by data type and ordered by frequency in this review

| Data type | Public dataset | Medical fields[1] | Relevant variables | Number of papers |
|---|---|---|---|---|
| Tabular Static | MIMIC-III, MIMIC-IV | Cardiology, CC, ED, Geriatrics, Informatics & Policy, Pharmacy, Renal | Age, Ethnicity/Race, Gender/Sex, Language, Marital status, Socioeconomics, Not Specified | 16 |
| Tabular Static | MEPS | Informatics & Policy, Oral Health, Public | Age, Ethnicity/Race, Gender/Sex, Socioeconomics, Not Specified | 11 |
| Tabular Static | Heritage Health | Informatics & Policy | Age, Gender/Sex, Not Specified | 8 |
| Tabular Static | SEER | Cancer, Gastroenterology, Hepatology, Pathology, Pulmonary | Age, Ethnicity/Race, Gender/Sex, Health status, Socioeconomics | 5 |
| Tabular Static | eICU | CC, Geriatrics | Age, Ethnicity/Race, Gender/Sex, Institute, Language | 4 |
| Tabular Static | Heart Disease | Cardiology | Age | 3 |
| Tabular Static | SUPPORT | Cancer, Cardiology, CC, Gastroenterology, Hepatology, ID, Neurology, Pulmonary | Age, Ethnicity/Race, Gender/Sex | 3 |
| Tabular Static | ABIDE | Mental, Neurology, Radiology | Age, Ethnicity/Race, Gender/Sex, Health status | 2 |
| Tabular Static | BRFSS | Informatics & Policy | Age, Ethnicity/Race, Gender/Sex | 2 |
| Tabular Static | Diabetic Hospital Dataset | Endocrinology, Informatics & Policy | Age, Ethnicity/Race, Gender/Sex, Not Specified | 2 |
| Tabular Static | MICD | Informatics & Policy | Gender/Sex, Not Specified | 2 |
| Tabular Static | NIS | Cancer, Cardiology | Ethnicity/Race, Location, Socioeconomics | 2 |
| Tabular Static | OAI | Orthopedics, Radiology, Rheumatology | Age, Anthropometry, Education, Ethnicity/Race, Gender/Sex, Socioeconomics | 2 |
| Tabular Static | STARR | Cardiology, CC, Informatics & Policy | Age, Ethnicity/Race, Gender/Sex, Socioeconomics | 2 |
| Tabular Static | TCGA | Cancer, Pathology | Ethnicity/Race | 2 |
| Tabular Static | Twitter | O & G | Ethnicity/Race | 2 |
| Tabular Static | UCI Diabetes | Endocrinology | Age, Not Specified | 2 |
| Tabular Static | UCI Drug | Mental | Ethnicity/Race, Gender/Sex | 2 |
| Tabular Static | ABCD | Neurology, Pediatrics, Radiology | Ethnicity/Race | 1 |
| Tabular Static | ADNI | Neurology | Ethnicity/Race, Gender/Sex | 1 |
| Tabular Static | Action to Control Cardiovascular Risk in Diabetes (ACCORD) | Cardiology, Endocrinology | Age, Education, Ethnicity/Race, Gender/Sex | 1 |
| Tabular Static | American College of Surgeons Trauma Quality Improvement Program | ED | Ethnicity/Race | 1 |
| Tabular Static | American Family Cohort | Informatics & Policy | Ethnicity/Race | 1 |
| Tabular Static | AmsterdamUM Cdb | CC, Geriatrics | Age, Ethnicity/Race, Gender/Sex | 1 |
| Tabular Static | Anti-hypertensive and Lipid-Lowering Treatment to Prevent Heart Attack Trial (ALLHAT) | Cardiology, Pharmacology | Age, Education, Ethnicity/Race, Gender/Sex | 1 |
| Tabular Static | Arkansas Clinical Data Repository | Informatics & Policy | Location | 1 |
| Tabular Static | BH | ID, Pulmonary | Ethnicity/Race, Institute | 1 |
| Tabular Static | CBSA | Mental | Ethnicity/Race | 1 |
| Tabular Static | CCHMC EHR | Informatics & Policy, Pediatrics, Surgery | Location | 1 |
| Tabular Static | CDC 500 Cities data | Endocrinology, Public | Age, Ethnicity/Race, Socioeconomics | 1 |
| Tabular Static | CHESS | CC, ID | Age, Ethnicity/Race, Gender/Sex | 1 |
| Tabular Static | CRDC survey | Mental | Ethnicity/Race | 1 |
| Tabular Static | China Migrants Dynamic Survey | Informatics & Policy | Socioeconomics | 1 |
| Tabular Static | CoRDaCo dataset | ID, Informatics & Policy | Age, Ethnicity/Race, Gender/Sex, Location | 1 |
| Tabular Static | Coswara | ID | Age, Ethnicity/Race, Gender/Sex, Health status, Language, Location, Time | 1 |
| Tabular Static | Digital Hand Atlas | Radiology | Age, Anthropometry, Ethnicity/Race, Gender/Sex | 1 |
| Tabular Static | Emory University Hospital System | ID, Public | Ethnicity/Race, Gender/Sex | 1 |
| Tabular Static | Enterprise data warehouse | CC | Ethnicity/Race, Health status | 1 |
| Tabular Static | FAERS | Pharmacy | Gender/Sex | 1 |
| Tabular Static | FDA medical devices database | Eye | Ethnicity/Race | 1 |
| Tabular Static | FLChain | Public | Age, Gender/Sex | 1 |
| Tabular Static | GapMap | Mental | Ethnicity/Race | 1 |
| Tabular Static | Geisinger | ID | Age, Ethnicity/Race, Gender/Sex, Socioeconomics | 1 |
| Tabular Static | Get With The Guidelines-Heart Failure | Cardiology, Informatics & Policy | Ethnicity/Race, Gender/Sex, Socioeconomics | 1 |
| Tabular Static | HECKTOR 2022 challenge dataset | Cancer, Otolaryngology, Radiology | Age, Gender/Sex, Health status, Others | 1 |
| Tabular Static | Health Facts database | Endocrinology | Gender/Sex | 1 |
| Tabular Static | Health and Retirement Study | Geriatrics, Informatics & Policy | Age, Ethnicity/Race, Gender/Sex, Location | 1 |
| Tabular Static | IBM MarketScan Medicaid Database | Mental, O & G | Ethnicity/Race | 1 |
| Tabular Static | IHDP | Pediatrics | Ethnicity/Race | 1 |
| Tabular Static | IWPC dataset | Pharmacy | Ethnicity/Race | 1 |
| Tabular Static | Indian Liver Patient Dataset (ILPD) | Hepatology | Gender/Sex | 1 |
| Tabular Static | Infections in Oxfordshire Research Database (IORD) | ED, ID | Ethnicity/Race, Institute | 1 |
| Tabular Static | MAIKI | Mental | Gender/Sex | 1 |
| Tabular Static | MMRF CoMMpass | Cancer, Pathology | Ethnicity/Race | 1 |
| Tabular Static | NPD-CRIS linkage data | Mental, Pediatrics | Ethnicity/Race, Language | 1 |
| Tabular Static | National Central Cancer Registry of China | Cancer, Gastroenterology | Gender/Sex, Location | 1 |
| Tabular Static | National Comorbidity Survey Replication Adolescent Supplement (NCS-A) | Mental | Ethnicity/Race, Gender/Sex | 1 |
| Tabular Static | Nursing Home Compare Star Ratings | Geriatrics, Public | Age, Ethnicity/Race, Gender/Sex, Socioeconomics | 1 |
| Tabular Static | OASIS | Neurology | Gender/Sex | 1 |
| Tabular Static | OUH | ID, Pulmonary | Ethnicity/Race, Institute | 1 |
| Tabular Static | Older Adults | Geriatrics | Gender/Sex | 1 |
| Tabular Static | Optum CDM | Cardiology, Informatics & Policy | Age, Gender/Sex | 1 |
| Tabular Static | Optum Labs Data Warehouse | Informatics & Policy | Ethnicity/Race | 1 |
| Tabular Static | Organ Procurement and Transplantation Network (OPTN) | Hepatology | Age, Ethnicity/Race, Gender/Sex | 1 |
| Tabular Static | Our world in Data | ID | Gender/Sex | 1 |
| Tabular Static | PHENOM | Neurology, Radiology | Age, Ethnicity/Race, Gender/Sex | 1 |
| Tabular Static | PRAp | Cancer, Urology | Ethnicity/Race, Socioeconomics | 1 |
| Tabular Static | PUH | ID, Pulmonary | Ethnicity/Race, Institute | 1 |
| Tabular Static | SPRINT | Cardiology | Ethnicity/Race, Gender/Sex | 1 |
| Tabular Static | Stanford Medicine Research Data Repository | Cardiology | Age, Ethnicity/Race, Gender/Sex | 1 |
| Tabular Static | Stanford-Colorado | Radiology | Age, Anthropometry, Ethnicity/Race, Gender/Sex | 1 |
| Tabular Static | Statistics NZ Integrated Data Infrastructure (IDI) | O & G | Age, Education, Ethnicity/Race, Location, Socioeconomics | 1 |
| Tabular Static | Systolic Blood Pressure Intervention Trial (SPRINT) | Cardiology | Age, Education, Ethnicity/Race, Gender/Sex | 1 |
| Tabular Static | TARGET | Cancer, Pathology | Ethnicity/Race | 1 |
| Tabular Static | The Global health 50/50 | ID | Gender/Sex | 1 |
| Tabular Static | The Standard Transparent Analysis and Research (STAR) organ transplant dataset | CC, Hepatology | Ethnicity/Race, Gender/Sex | 1 |
| Tabular Static | UCI ASD | Mental | Gender/Sex | 1 |
| Tabular Static | UCI Cleveland heart dataset | Cardiology | Gender/Sex | 1 |
| Tabular Static | UCI Heart Health | Cardiology | Age, Gender/Sex | 1 |
| Tabular Static | UCI Obesity | Public | Gender/Sex | 1 |
| Tabular Static | UCLA Consortium dataset | Mental, Radiology | Gender/Sex, Health status | 1 |
| Tabular Static | UHB | ID, Pulmonary | Ethnicity/Race, Institute | 1 |
| Tabular Static | UK Biobank | Endocrinology | Gender/Sex, Others, Socioeconomics | 1 |
| Tabular Static | UK Biobank polygenic | Genetics | Ethnicity/Race | 1 |
| Tabular Static | United Nation database | O & G, Pediatrics | Clinician expertise, Gender/Sex, Location | 1 |
| Tabular Static | Victorian minimum emergency dataset (VEMD) | ED | Gender/Sex, Location | 1 |
| Tabular Static | iSTAGING | Neurology, Radiology | Age, Ethnicity/Race, Gender/Sex | 1 |
| Tabular Static | rRNA gene sequences datasets | ID, O & G, Pathodology | Ethnicity/Race | 1 |
| Tabular Static | the OCT equipment user manual of the following contemporary OCT equipment | Eye | Ethnicity/Race | 1 |
| Tabular Temporal | MIMIC-III, MIMIC-IV | Cancer, CC, ID, Informatics & Policy, Pulmonary | Age, Ethnicity/Race | 3 |
| Tabular Temporal | SEER | Cancer, Informatics & Policy, Pulmonary | Age, Ethnicity/Race | 2 |
| Tabular Temporal | 2019-nCoV | ED, ID | Not Specified | 1 |
| Tabular Temporal | ACS 5-year | Public | Ethnicity/Race, Socioeconomics | 1 |
| Tabular Temporal | ASD | Mental, Pediatrics | Ethnicity/Race, Gender/Sex | 1 |
| Tabular Temporal | ATUS | Public | Age, Gender/Sex | 1 |
| Tabular Temporal | Enterprise data warehouse | CC | Ethnicity/Race, Health status | 1 |
| Tabular Temporal | Inmate Mental Health | Mental | Ethnicity/Race | 1 |
| Tabular Temporal | MEPS | Informatics & Policy | Ethnicity/Race | 1 |
| Tabular Temporal | MIDRC | ID | Age, Ethnicity/Race, Gender/Sex | 1 |
| Tabular Temporal | NCDB | Urology | Ethnicity/Race | 1 |
| Tabular Temporal | National Health and Nutrition Examination Survey | Public | Age | 1 |
| Tabular Temporal | Northern Alberta Cancer Dataset (NACD) | Neurology | Not Specified | 1 |
| Tabular Temporal | Optum‚Äôs de-identified EHR and Clinformatics data Mart database | Cancer, Pulmonary | Ethnicity/Race | 1 |
| Tabular Temporal | UK Biobank data | Cardiology, Endocrinology, Pathodology | Age, Ethnicity/Race, Gender/Sex | 1 |
| Tabular Temporal | World bank Life expectancy vs. healthcare expenditure | Public | Not Specified | 1 |
| Tabular Temporal | World bank data | O & G, Pediatrics | Gender/Sex | 1 |
| Image | MIMIC-CXR | Cardiology, Dermatology, Pulmonary, Radiology | Age, Ethnicity/Race, Gender/Sex, Skin tone, Socioeconomics | 6 |
| Image | CheXpert | Cardiology, Dermatology, Pulmonary, Radiology | Age, Ethnicity/Race, Gender/Sex, Skin tone | 5 |
| Image | Chest-Xray8, Chest-Xray14 | Cardiology, Pulmonary, Radiology | Age, Ethnicity/Race, Gender/Sex | 5 |
| Image | Fitzpatrick17k | Cancer, Dermatology | Age, Gender/Sex, Health status, Skin tone | 5 |
| Image | ISIC | Cancer, Dermatology, Radiology | Gender/Sex, Skin tone | 5 |
| Image | ABIDE | Mental, Neurology, Radiology | Age, Ethnicity/Race, Gender/Sex, Health status, Others | 3 |
| Image | EyePACS | Endocrinology, Eye | Skin tone, Not Specified | 3 |
| Image | TCGA | Cancer, Pathodology, Pathology, Pulmonary, Renal | Ethnicity/Race, Institute, Not Specified | 3 |
| Image | UK Biobank CMR images | Cardiology, Radiology | Ethnicity/Race, Gender/Sex | 3 |
| Image | ABCD | Neurology, Pediatrics, Radiology | Ethnicity/Race | 2 |
| Image | DDI | Cancer, Dermatology, ID, Pulmonary, Radiology | Age, Ethnicity/Race, Skin tone | 2 |
| Image | OAI | Orthopedics, Radiology, Rheumatology | Education, Ethnicity/Race, Gender/Sex, Socioeconomics | 2 |
| Image | PathMNIST | Cancer, Gastroenterology, Pathology | Not Specified | 2 |
| Image | APTOS | Endocrinology, Eye | Not Specified | 1 |
| Image | AREDS | Eye, Geriatrics | Age, Ethnicity/Race, Gender/Sex | 1 |
| Image | Alzheimer‚Äôs Disease Neuroimaging Initiative | Neurology, Radiology | Gender/Sex | 1 |
| Image | BCN20000 | Cancer, Dermatology | Skin tone | 1 |
| Image | CXR | Cancer, Dermatology, ID, Pulmonary, Radiology | Age, Ethnicity/Race | 1 |
| Image | Covid19 | ID | Not Specified | 1 |
| Image | Dermnet | Cancer, Dermatology | Skin tone | 1 |
| Image | Emory-CXR | Pulmonary, Radiology | Age, Ethnicity/Race, Gender/Sex | 1 |
| Image | Guangzhou pediatric dataset | Pediatrics, Pulmonary, Radiology | Age | 1 |
| Image | HAM10000 | Cancer, Dermatology | Skin tone | 1 |
| Image | HCP | Neurology, Radiology | Ethnicity/Race | 1 |
| Image | HECKTOR 2022 challenge dataset | Cancer, Otolaryngology, Radiology | Age, Gender/Sex, Health status, Others | 1 |
| Image | Human Connectome Project | Neurology | Gender/Sex | 1 |
| Image | International Skin Imaging Collaboration 2019 | Cancer, Dermatology | Age, Gender/Sex, Health status | 1 |
| Image | Kaggle 2020 Melanoma Competition dataset | Dermatology | Skin tone | 1 |
| Image | MIDRC | Pulmonary, Radiology | Age, Ethnicity/Race, Gender/Sex | 1 |
| Image | NCANDA | Neurology, Pediatrics, Radiology | Not Specified | 1 |
| Image | NIH Chest X-Ray | Pulmonary, Radiology | Gender/Sex | 1 |
| Image | National Genome Research Network Plus | Mental | Age, Gender/Sex | 1 |
| Image | OASIS | Neurology | Gender/Sex | 1 |
| Image | OHTS | Eye | Age, Ethnicity/Race, Gender/Sex | 1 |
| Image | PHENOM | Neurology, Radiology | Age, Ethnicity/Race, Gender/Sex | 1 |
| Image | RSNA RIC | Pediatrics, Radiology | Not Specified | 1 |
| Image | Shinshu set | Cancer, Dermatology | Skin tone | 1 |
| Image | The Mobile Ocular Biometrics In Unconstrained Settings | Eye | Ethnicity/Race, Others | 1 |
| Image | The Multi-Angle Sclera Dataset | Eye | Ethnicity/Race, Others | 1 |
| Image | The Sclera Blood Vessels Periocular and Iris | Eye | Ethnicity/Race, Others | 1 |
| Image | The Sclera Liveness Dataset | Eye | Ethnicity/Race, Others | 1 |
| Image | The Sclera Mobile Dataset | Eye | Ethnicity/Race, Others | 1 |
| Image | UCLA Consortium dataset | Mental, Radiology | Gender/Sex, Health status | 1 |
| Image | chest-xray-pneumonia | Pulmonary | Not Specified | 1 |
| Image | iSTAGING | Neurology, Radiology | Age, Ethnicity/Race, Gender/Sex | 1 |
| Video | UBFC-RPPG | Informatics & Policy | Skin tone | 1 |
| Video | UCLA-rPPG | Hematology | Skin tone | 1 |
| Video | VITAL | Informatics & Policy | Skin tone | 1 |
| Signal | 1000 fragments | Cardiology | Age, Gender/Sex | 1 |
| Signal | 2021 PhysioNet/Computing in Cardiology Challenge | Cardiology | Age, Ethnicity/Race, Gender/Sex | 1 |
| Signal | Clinical TUH Abnormal EEG Corpus | Neurology | Gender/Sex | 1 |
| Signal | MIT BIH Arrhythmia database | Cardiology | Age, Gender/Sex | 1 |
| Signal | Tiles-2018 | Mental | Age, Ethnicity/Race, Gender/Sex, Institute, Language, Location, Others, Socioeconomics | 1 |
| Signal | UBFC-RPPG | Informatics & Policy | Skin tone | 1 |
| Signal | VITAL | Informatics & Policy | Skin tone | 1 |
| Audio | ADReSS challenge dataset | Neurology | Not Specified | 1 |
| Audio | DAIC-WOZ | Mental | Gender/Sex | 1 |
| Text | Twitter | Mental, O & G | Ethnicity/Race, Location | 3 |
| Text | MIMIC-III, MIMIC-IV | CC | Gender/Sex, Not Specified | 2 |
| Text | AURIN-PHA | Mental | Location | 1 |
| Text | FluTrack | ID, Public | Ethnicity/Race | 1 |
| Text | FluVacc | ID, Public | Ethnicity/Race | 1 |
| Text | Medical transcription dataset(mtsamples.com) | Informatics & Policy | Gender/Sex | 1 |
| Text | U.S. Social Security; U.S. Census | Informatics & Policy | Ethnicity/Race, Gender/Sex, Others, Time | 1 |

[1]refers to those medical fields where the databases were applied as observed in this review, rather than how these databases are defined

**Supplementary Table S2.** Distribution of AI research among medical fields as documented by published systematic reviews

| Medical fields | Number of screened AI research | Databases | Review | Search up to year | Note[1] |
|---|---|---|---|---|---|
| Informatics & Policy | / | / | / | / | |
| Cancer | 3391 | Medline | Suero-Abreu et al., 2022 | 2019 | |
| Radiology | 11209 | Web of Science | Kocak et al., 2023 | (2000-) 2021 | |
| Cardiology | 1172 | Medline | Suero-Abreu et al., 2022 | 2019 | |
| ID (Infectious Disease) | 3261 | Web of Science, Medline, Springer, IEEE, ScienceDirect | Alqaissi et al., 2022 | (2015-) 2022 | |
| Mental Health | 300 | PsycInfo, Cochrane, Medline, IEEE, ACM | Shatte et al., 2019 | 2018 | |
| CC (Critical Care) | 494 | Embase, Medline, Web of science, Cochrane, Google scholar | Van de Sande et al., 2021 | 2020 | |
| Neurology | 154 | Medline, Scopus, Web of Science | Segato et al., 2020 | 2020 | Chronic Obstructive Lung Disease |
| Pulmonology | 156 | Medline, DBLP | Exarchos et al., 2021 | 2020 | |
| Public Health | / | / | / | / | |
| Pediatric | 363 | Medline, Cochrane, the Cumulative Index to Nursing and Allied Health Literature Plus, Web of Science, EBSCO Dentistry & Oral Science Source | Hoodbhoy et al., 2021 | 2020 | |
| Dermatology | 64 | Medline | Thomsen et al., 2019 | 2018 | |
| ED (Emergency Department) | 395 | Medline-OVID, Embase, CINAHL, IEEE | Kirubarajan et al., 2020 | 2020 | |
| Endocrinology | 17 | Medline | Giorgini et al., 2024 | 2023 | |
| O & G (Obstetrics and Gynecology) | 66 | Medline | Dhombres et al., 2022 | 2020 | |
| Eye | 69 | Medline, Scopus | Nuzzi et al., 2021 | 2021 | |
| Gastroenterology | 73 | Medline, CINAHL, Cochrane, Web of Science | Parkash et al., 2022 | 2021 | |
| Geriatrics | 105 | Medline, EBASE, Cochrane, Web of Science, PyscInfo, | Ma et al., 2023 | 2023 | |
| | | CNKI, SinoMed, WANFANG, VIP | | | |
| Hepatology | 150 | Medline | Balsano et al., 2022 | 2022 | |
| Pharmacy | 43 | Medline, Google Scholar, Scopus | Chalasani et al., 2023 | 2023 | |
| Surgery | / | / | Varghese et al., 2024 | 2024 | number was not reported |
| Orthopedics | 223 | Embase, Medline, Scopus | Federer et al., 2021 | 2023 | |
| Pathology | 100 | Medline, Embase, CENTRAL | McGenity et al., 2024 | 2024 | |
| Urology | 112 | / | Chen et al., 2022 | 2020 | database was not reported |
| Otolaryngology | 54 | Medline | Bur et al., 2019 | 2019 | |
| Family Medicine | 405 | Medline-OVID, Embase, CINAHL, Cochrane, Web of Science, Scopus, IEEE, ACM Digital Library, MathSciNet, AAAI | Kueper et al., 2020 | 2020 | |
| Renal (Nephrology) | 218 | the Institute for Scientific Information Web of Knowledge database | Park et al., 2021 | 2020 | |
| Rheumatology | 91 | Medline, Scopus, Web of Science, Rheumatology journals | Madrid-García et al., 2023 | 2022 | |
| Anesthesiology | 173 | Medline, Embase, Web of Science, IEEE | Hashimoto et al., 2020 | 2018 | |
| Hematology | 53 | Medline | Kotsyfakis et al., 2022 | 2022 | Hematology oncology |
| Oral Health | 178 | Cochrane, Medline, Scopus, IEEE | Morch et al. 2021 | 2020 | |
| Immunology | 169 | Medline, Embase | Stafford et al., 2020 | 2018 | Autoimmune disease |
| Occupational Medicine | 260 | Medline, Google Scholar, Scopus | Pishgar et al, 2021 | 2019 | |
| PM & R (Physical Medicine and Rehabilitation) | 29 | Medline, Embase, CINAHL, NARIC, Web of Science, OpenGrey | Sumner et al, 2023 | 2021 | |

[1]When no systematic review exists for a specific medical field due to its broad nature, an alternative smaller field was used.

**Supplementary Table S3.** Search strategies

| Database | Search Strategy |
|---|---|
| MEDLINE | (("machine learning") OR ("deep learning") OR ("artificial intelligence")) AND ("fair*"[tiab] OR "equality"[tiab] OR "equity"[tiab] OR ((“alleviat*”[tiab] OR"mitigat*"[tiab] OR "reduc*"[tiab] OR "eliminat*"[tiab] OR "evaluat*"[tiab] OR "assess*"[tiab] OR "measure*"[tiab]OR "detect*"[tiab]) NEAR ("bias*"[tiab] OR "disparit*"[tiab] OR "inequality"[tiab] OR “inequity”[tiab]))) AND (("health*") OR ("medicine") OR ("clinical") OR ("medical")) |
| Web of Science | ALL=(("machine learning") OR ("deep learning") OR ("artificial intelligence")) AND AB=("fair*” OR “equality” OR “equity” OR ((“alleviat*” OR “mitigat*” OR  "reduc*" OR “eliminat*” OR “evaluat*” OR “assess*” OR “measure*” OR “detect*”) NEAR (“bias*” OR “disparit*” OR “inequality” OR “inequity”))) AND ALL=(("health*") OR ("medicine") OR ("clinical") OR ("medical")) |
| Embase | ('machine learning' OR 'deep learning' OR 'artificial intelligence') AND (‘fair*’:ab,ti OR ‘equality’:ab,ti OR ‘equity’:ab,ti OR ((‘alleviat*’ OR ‘mitigat*’ OR ‘reduc*’ OR ‘eliminat*’ OR ‘evaluat*’ OR ‘assess*’ OR ‘measure*’ OR ‘detect*’) NEAR/15 (‘bias*’ OR ‘disparit*’ OR ‘inequality’ OR ‘inequity’)):ab,ti) AND ('health*' OR 'medicine' OR 'clinical' OR 'medical') |
| IEEE Xplore | ((("Full Text & Metadata":"machine learning" OR "Full Text & Metadata":"deep learning" OR "Full Text & Metadata":"artificial intelligence") AND ("Abstract":"fair" OR "Abstract":"fairness" OR "Abstract":"fairly" OR "Abstract":"equality" OR "Abstract":"equity" OR (("Abstract":“alleviat*” OR "Abstract":"mitigat*" OR "Abstract":"reduc*" OR "Abstract":"eliminat*" OR "Abstract":"evaluat*" OR "Abstract":"assess*" OR "Abstract":"measure*"OR "Abstract":"detect*") NEAR ("Abstract":"bias" OR “Abstract”:”biases” OR "Abstract":"disparity" OR “Abstract”:”disparities” OR "Abstract":"inequality" OR “Abstract”:”inequity”))) AND ("Full |
| | Text & Metadata":"health*" OR "Full Text & Metadata":"medicine" OR "Full Text & Metadata":"clinical" OR "Full Text & Metadata":"medical"))) |
| ACM Library | [[All: "machine learning"] OR [All: "deep learning"] OR [All: "artificial intelligence"]] AND [[Abstract: "fair*"] OR [Abstract: "equality"] OR [Abstract: "equity"] OR [[[Abstract: "alleviat*"] OR [Abstract: "mitigat*"] OR [Abstract: "reduc*"] OR [Abstract: "eliminat*"] OR [Abstract: "evaluat*"] OR [Abstract: "assess*"] OR [Abstract: "measure*"] OR [Abstract: "detect*"]] AND [[Abstract: "bias*"] OR [Abstract: "disparit*"] OR [Abstract: "inequality"]]]] AND [[All: "health*"] OR [All: "medicine"] OR [All: "clinical"] OR [All: "medical"]] |